\documentclass{egpubl}
\usepackage{STAG2025}

\WsPaper           

\usepackage[T1]{fontenc}
\usepackage{dfadobe}  

\usepackage{cite}

\BibtexOrBiblatex

\electronicVersion
\PrintedOrElectronic

\ifpdf \usepackage[pdftex]{graphicx} \pdfcompresslevel=9
\else \usepackage[dvips]{graphicx} \fi

\usepackage{egweblnk} 

\title[ReCoGS]{ReCoGS: Real-time ReColoring for Gaussian Splatting scenes}

\author[Lorenzo Rutayisire \& Nicola Capodieci \& Fabio Pellacini]
{\parbox{\textwidth}{
    \centering
    Lorenzo Rutayisire \quad
    Nicola Capodieci \quad
    Fabio Pellacini
} \\
{\parbox{\textwidth}{\centering University of Modena and Reggio Emilia}
}}

\newcommand{\quotes}[1]{``#1''}

\usepackage{amsmath,mathtools}
\usepackage{amssymb}
\usepackage{array}
\usepackage{booktabs}
\usepackage{cleveref}
\usepackage{dblfloatfix}
\usepackage{float}
\usepackage{graphicx}
\usepackage{makecell}
\usepackage{placeins}
\usepackage{subcaption}

\newcolumntype{P}[1]{>{\centering\arraybackslash}p{#1}}

\DeclareMathOperator{\sign}{sign}

\begin{document}


\maketitle
\begin{abstract}
Gaussian Splatting has emerged as a leading method for novel view synthesis, offering superior training efficiency and real-time inference compared to NeRF approaches, while still delivering high-quality reconstructions. Beyond view synthesis, this 3D representation has also been explored for editing tasks. Many existing methods leverage 2D diffusion models to generate multi-view datasets for training, but they often suffer from limitations such as view inconsistencies, lack of fine-grained control, and high computational demand. In this work, we focus specifically on the editing task of recoloring. We introduce a user-friendly pipeline that enables precise selection and recoloring of regions within a pre-trained Gaussian Splatting scene. To demonstrate the real-time performance of our method, we also present an interactive tool that allows users to experiment with the pipeline in practice. Code is available at \href{https://github.com/loryruta/recogs}{https://github.com/loryruta/recogs}.

\end{abstract}  

\section{Introduction}
\label{sec:intro}

3D Gaussian Splatting (3DGS) \cite{3dgs} represents the state-of-the-art method for novel view synthesis,
outperforming Neural Radiance Field methods (NeRF) \cite{nerf}.
Specifically, 3DGS offers faster training time, yet achieving a reconstruction quality comparable to the most recently presented NeRF methods, enabling real-time performance on customer-grade hardware.
In NeRF-based approaches the outgoing radiance along a viewing direction is encoded in a Multi-Layer Perceptron (MLP) network, whereas
Gaussian Splatting representation is more explicit and consists of a set of 3D gaussians positioned to match the training dataset.

Despite being more explicit,
the 3DGS representation remains challenging for editing tasks, primarily because 3D gaussians do not represent objects geometry.
In the original work,
gaussians are optimized with only a photometric loss, which allows them to be placed arbitrarily in space as long as the accumulated radiance matches the ground-truth.
However, since most gaussians are initialized at precise 3D keypoints, and since --- to account for view-consistency constraints --- the optimizer is implicitly encouraged
to move them towards actual object surfaces, it is still plausible to assume that the final set of gaussians approximates the scene's geometry.

This assumption is leveraged by editors like \cite{splatshop},
\cite{supersplat}, that propose a manual selection of the gaussians, thus enabling various editing operations, including tinting.
In these tools, tinting is applied uniformly to the selected gaussians.
However, due to the manual selection of the gaussians, and the view-independence of the applied tint,
we observe that precise edit application is cumbersome and color bleeding in unwanted scene regions is not uncommon.

In order to facilitate the selection of editing regions,
previously published approaches are based on the idea of embedding segmentation labels within gaussians \cite{saga}, \cite{gaussian_grouping}, \cite{gaussian_editor},
or language features for open-vocabulary 3D segmentation \cite{langsplat}.
However, these methods require costly re-training of the 3DGS scene, making them unsuitable for in-place editing.
Additionally, the granularity of the selections is inherently limited by the chosen segmentation model (e.g. SAM \cite{sam}).

In this context, another worthy approach to editing is represented by mesh extraction.
The idea is that if it is possible to extract a triangulated mesh from the 3DGS representation,
then it is possible to edit it with any traditional 3D editing software.
Notable meshing works are \cite{sugar}, \cite{gaussian_frosting}, \cite{2dgs}, \cite{gs2mesh}, while \cite{texturegs} only focuses on texture extraction.
Additionally, \cite{gsops} is a plugin for Houdini providing connection between 3DGS and mesh representations.
All the aforementioned approaches require re-training,
which is generally more costly than the original 3DGS training as additional geometric constraints are introduced to align gaussians to surfaces \cite{sugar}, \cite{2dgs}, 
often degrading reconstruction quality.

Modern editing research focuses on leveraging the vast knowledge of 2D diffusion models to generate or edit 3D content, given a textual or image prompt.
A milestone work in this direction is \cite{instructnerf2nerf}, in which the authors propose to perform iterative dataset updates (using \cite{instructpix2pix}) and to converge the 3D scene towards them. This method was tailored for NeRF, follow-up work proposed a similar technique for 3DGS scenes \cite{instructgs2gs}, \cite{gaussian_editor}.
One major issue of the aforementioned methods is that dataset images are updated independently, leading the scene to be optimized with noisy gradients due to inconsistent views.
Several recent works \cite{vcedit}, \cite{gaussctrl}, \cite{dge}, \cite{splatflow}, \cite{editsplat} propose different solutions to mitigate view inconsistencies,
but they generally fail at producing pixel-perfect results. This is usually due to the diffusion model operating in a low-resolution latent space
which leads the final edited scene to have blurry regions or still showcase inconsistencies.  
Moreover, the textual prompt selection neglect fine-grain edits, although usually the target region can be constrained with a bounding box.


Our recoloring pipeline, to the best of our knowledge, stands out from all the prior work.
We propose a novel selection method that allows the user to paint a 2D selection/mask on a reference view at pixel resolution.
We then propose a method for unprojecting such 2D selection to a 3D mask exploiting a pre-trained depth estimation model.
Such 3D selection is then re-projected to training views onto which the edit is applied.
The pre-trained Gaussian Splatting scene is fine-tuned towards the edited dataset in background,
leveraging an optimizer implementation tailored to our problem.
Considering the pipeline as a whole, our approach showcase great interactivity to fine-grain recolor anything in a Gaussian Splatting scene.
We provide the code for a demonstrative editor implementing our method at [this URL].


\section{Related Work}
\label{sec:related_work}

Most of the literature on Gaussian Splatting editing aims at leveraging 2D diffusion models
to generate or edit 3D content (e.g. \cite{stable_diffusion_2}, \cite{instructpix2pix}).
Hence, recoloring becomes one of the possibly many editing operations available.

Coupling Gaussian Splatting training with diffusion models carries a significant additional computation cost that makes these methods unsuitable for real-time editing.


\subsection{Non Multi-View Consistent Diffusion}

A milestone work in 3D editing of NeRF is InstructNerf2Nerf \cite{instructnerf2nerf},
where they propose to update training images using a text-prompted 2D diffusion model \cite{instructpix2pix},
and alternate these updates with optimization iterations of the 3D scene.
The adaptation of this work to 3DGS is Instruct-GS2GS \cite{instructgs2gs}. 

The drawback of these approaches is that dataset images are updated independently, which may lead
the diffusion model to generate view-inconsistent outputs, hindering the optimization of the 3D scene. 

In TIP-Editor \cite{tip_editor}, authors leveraged a 2D diffusion model \cite{stable_diffusion_2} which can be fed with text and image prompt, and a bounding box.
The bounding box defines where the edit is placed and a localization loss is introduced to apply the edit in the target area.
Although this method provides an innovative way of selecting the region of interest, hence providing additional granularity for the edits,
it still suffers from the same issue of inconsistency among views. Such issues have been addressed in more recent methodologies. 


\subsection{Multi-View Consistent Diffusion}

In VcEdit \cite{vcedit}, authors extended an existing 2D diffusion model \cite{infedit} with two additional modules to achieve multi-view consistency.
The model's input are multiple views of the area to be edited and a text prompt.
The first module is used to ensure that attention maps from multiple views are consistent in 3D space,
while the second is an intermediate step of decoding/re-encoding of the multi-view latents to further strengthen their consistency.

GaussCtrl \cite{gaussctrl} proposes to condition the diffusion model  \cite{controlnet} with approximate depth maps extracted from the gaussians position.
To unify individual edits of images, another module is employed to cross-attend latents from different views.

DGE \cite{dge} treats adjacent training images as a sequence, forming a video.
Then, techniques are employed from video editing to consistently edit all the frames.
Specifically, they introduce a spatio-temporal attention,
and they propagate edit features to multiple frames leveraging the epipolar constraints. 



EditSplat \cite{editsplat} proposes a method to address multi-view consistency and optimization efficiency.
They condition the diffusion model with multi-view informed images, created by re-projecting edited pixels across views, leveraging the 3DGS depth map.
Additionally, they employ attention maps from the diffusion model to guide the optimization and enhance its efficiency.
Regarding the runtime, they report 6 minutes on a RTX A6000 GPU, suggesting their method not to be interactive.


\subsection{ReTexturing and ReColoring}

To the best of our knowledge, ICE-G \cite{iceg} and PaintSplat \cite{paintsplat} are the literature works closer to our recoloring task.
They both address re-texturing and re-coloring, without relying on 2D diffusion models.

In ICE-G \cite{iceg} authors propose a pipeline to select and edit objects' texture or color.
Their selection method consists in segmenting the training images using SAM \cite{sam},
and then, once the user selects a region intended to edit,
the same region is found in other views by extracting and matching DINO features.
Later, the texture or color is applied in multiple views and the scene then retrained to converge towards the edited dataset.
Although no code is provided, they claim that their method runs in 21 min on a NVIDIA A40 GPU, indicating it is unlikely to reach real-time interactive performance on customer grade devices.

Ultimately, PaintSplat \cite{paintsplat} is a method for painting a 3D stamp onto a pre-trained 3DGS scene.
The 3D stamp is a set of gaussians that can be prepared using segmentation or box selection.
The proposed method consists of placing and deforming the 3D stamp, to make contiguous sequence along the user stroke (i.e. a 3D spline) and eventually refining the edit with a diffusion model.
PaintSplat covers re-texturing and re-coloring, authors claim their method to run within real-time latencies;
optimization, typically the most time-consuming stage, solely occurs to refine the edit and is performed in background.


\section{Method}

\subsection{Preliminaries}


In Gaussian Splatting, the scene is represented as a set of 3D translucent gaussians, that are optimized through a differentiable rendering pipeline to match the target images. Each gaussian geometry is described by a covariance matrix, constructed as a combination of a rotation and a scaling matrix to ensure positive definiteness during the optimization process:
\begin{equation}
\Sigma = \mathbf{R} \mathbf{S} \mathbf{S}^T \mathbf{R}^T
\end{equation}

Every gaussian is also assigned with an opacity $\sigma$ and a view-dependent color $\mathbf{f}(\theta, \phi)$, which is parameterized by the spherical harmonics coefficients $\mathbf{C}_l^m$. 
\begin{equation}
\mathbf{f}(\theta, \phi) = \sum_{l = 0}^3 \sum^{l}_{m = -l} \mathbf{C}_l^m Y_l^m (\theta, \phi)
\end{equation}

During rendering, the color of each pixel is determined by integrating contributions from the gaussians along the viewing ray and accumulating their colors with alpha blending:
\begin{equation} \label{eq:gs_blending}
\begin{gathered}
\mathbf{y} = 
\sum^N_{i = 1}
\mathbf{f}_i(\mathbf{\theta, \phi}) \cdot \sigma_i \cdot
\mathcal{G}_i \cdot T_{i}\\
T_1 = 1\\
T_{i + 1} = T_i \cdot (1 - \alpha_i)\\
\end{gathered}
\end{equation}

In the original gaussian splatting paper the scene is optimized with a photometric loss, therefore the gaussians are arranged to match the visual appearance rather than the geometry.
\begin{equation} \label{eq:gs_loss}
\begin{gathered}
\mathcal{L} = (1 - \lambda) \mathcal{L}_1 + \lambda (1 - \mathcal{L}_{SSIM})
\end{gathered}
\end{equation}

In this scenario, we have developed an interactive selection method, which enables a user to select regions of the scene and to apply an \quotes{edit} to them. For the sake of simplicity, we chose the edit to be tinting (or recoloring).

\subsection{Overview}

\begin{figure*}[!tb]
    \centering
    \begin{subfigure}{0.19\textwidth}
        \includegraphics[width=\linewidth]{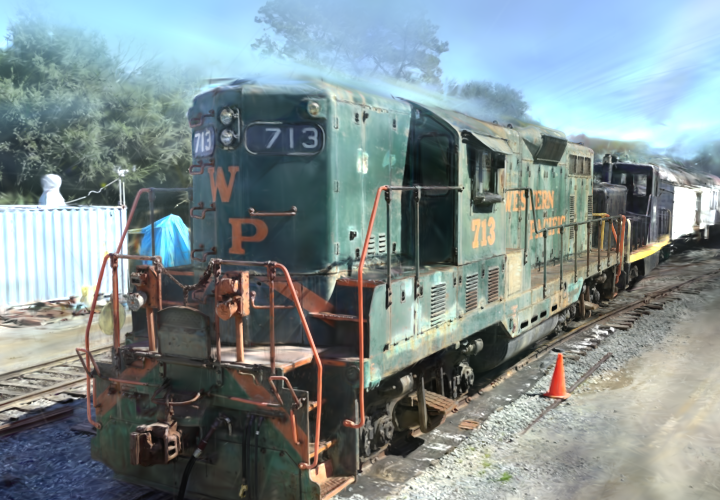}
        \caption{}
    \end{subfigure}
    \begin{subfigure}{0.19\textwidth}
        \includegraphics[width=\linewidth]{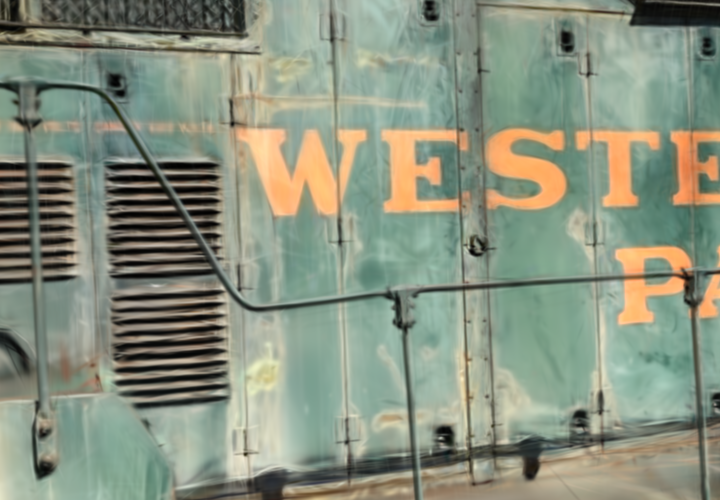}
        \caption{}
    \end{subfigure}
    \begin{subfigure}{0.19\textwidth}
        \includegraphics[width=\linewidth]{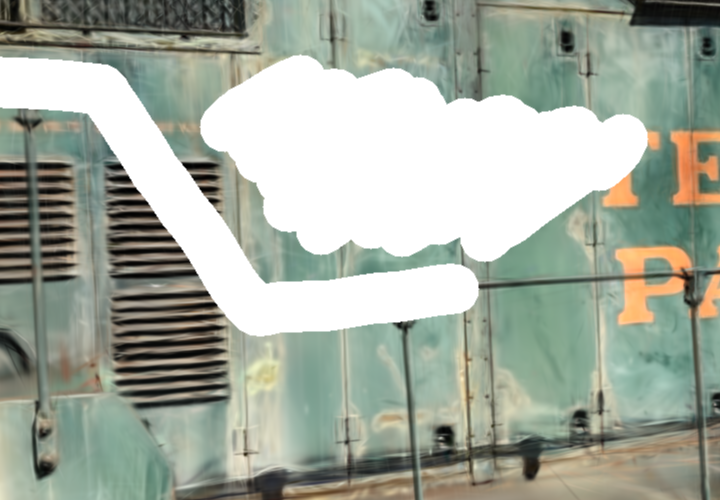}
        \caption{}
    \end{subfigure}
    \begin{subfigure}{0.19\textwidth}
        \includegraphics[width=\linewidth]{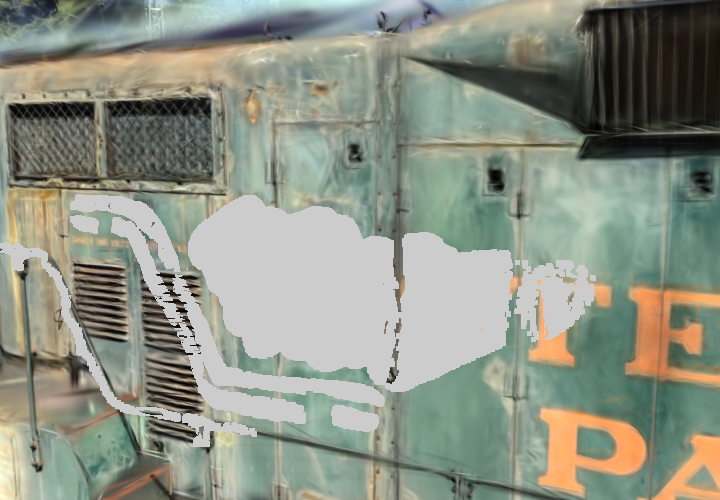}
        \caption{}
    \end{subfigure}
    \begin{subfigure}{0.19\textwidth}
        \includegraphics[width=\linewidth]{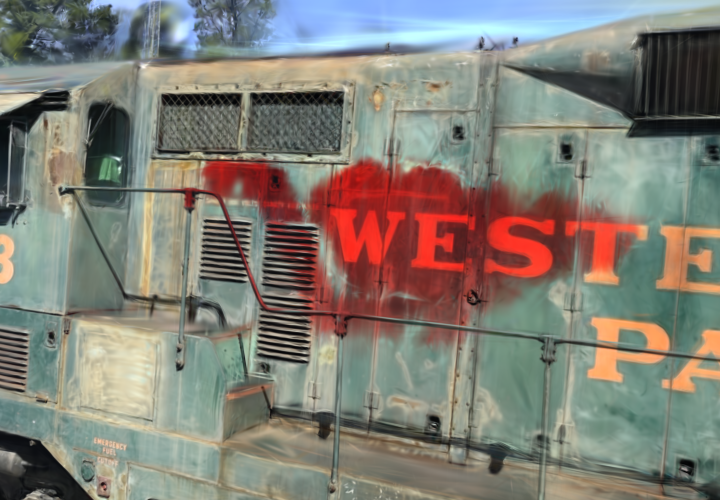}
        \caption{}
    \end{subfigure}
    \caption{ReCoGS editing pipeline. The user visits the scene through our editor (a), stops at an arbitrary view (b), selects pixels intended to edit (c), pixels are unprojected to a 3D pointcloud (d), the editing is applied in background (e).}
    \label{fig:recogs_pipeline}
\end{figure*}

We propose an interactive editing pipeline, that we have implemented in our editing tool ReCoGS \url{https://github.com/loryruta/ReCoGS}. The pipeline is outlined in the following steps (refer to \cref{fig:recogs_pipeline}):

\begin{enumerate}
\item The scene is first loaded into the editor and the user is given with the possibility to explore it.
\item The user stops at one view and draw a 2D selection mask.
\item The 2D selection is unprojected into a 3D pointcloud.
\item The pointcloud is projected onto the training views to provide a 2D mask on each training view to apply the edit.
\item The edit operation (i.e., recoloring) is applied on the pixels marked by the projected selection mask.
\item The scene parameters are optimized in background using the edited training views as a ground truth.
\item The user is given with the possibility to continuously edit the selection mask, thereby drifting the training objective.
\end{enumerate}

\subsubsection{2D Selection} \label{2d_selection}

When the user enters the selection mode, its current viewpoint is captured and a UI is opened to allow them to zoom and drag it, as if it was an image editor. Eventually the user can select image pixels the brush tool, and clear the selection with the rubber tool.

Once the user is satisfied with the selection, they exit the selection mode and the 2D mask is unprojected to a 3D pointcloud. To perform the unprojection, the camera parameters (both intrinsic and extrinsic) and the  depthmap need to be known. The former are already provided, and we use them to render the scene within the tool, but the depth map is unknown.

\subsubsection{Predicting the Depth map}

A rough depth map can be obtained by taking a \quotes{median} of the gaussians depth: when the transmittance $T$ becomes lower than a threshold $\tau$, heuristically make the last gaussian's $z$ the pixel's depth. Although this depth map can be computed efficiently, it turned out not to represent the real depth in correspondence of flat texture-less regions and very thin objects (e.g. poles). 

After trying several depth estimation methods belonging to diverse categories (e.g., \cite{depth_anything_v2}, \cite{dlnr}, \cite{pcvnet}), such as Monocular Depth Estimation, Stereo Matching and Multi-View Stereo; we have found PCVNet \cite{pcvnet} to be the one which worked best in terms of accuracy and runtime.

PCVNet is a deep model for stereo matching: the input is a pair of images (the \textit{left} and the \textit{right} image), and the output is a \textit{disparity map} $s$ from which the depth map $d$ can be computed knowing the view-space shift $\beta$ and focal length $f_x$ for every pixel $p$ as:
\begin{equation} \label{eq:disparity_to_depth}
d_p = f_x \dfrac{\beta}{s_p}
\end{equation}

As we are rendering a synthetic scene, we have access to an unlimited number of views which would mean a greater source of information for the depth prediction. Following this rationale, Multi-View Stereo models should be the ideal; however, we have found them to be resource intensive for customer-level hardware.
We still managed to exploit this insight by performing stereo matching both horizontally and vertically (what we call \quotes{Stereo-HV}) and aggregating the two depth maps $d'$ and $d''$ as $d = min(d', d'')$.

\subsubsection{3D Selection and Refinement}

\begin{figure}[htbp]
  \centering
  \includegraphics[width=\linewidth]{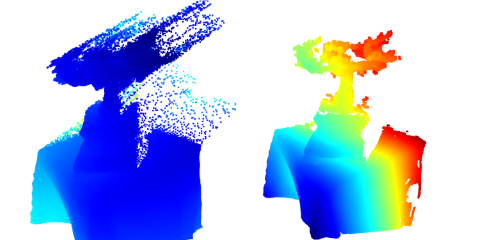}
  \caption{After unprojecting the 2D selection mask (left), we have found helpful to remove statistical outliers (right); that is, to remove points that are \quotes{more isolated} than other points. In our implementation we calculate the average distance with $16$ neighbors, and use a scale factor for standard deviation threshold of $0.007$.}
  \label{fig:pc_filtering}
\end{figure}

Provided with the 2D selection, camera parameters and the depth map, it is now possible to perform unprojection to obtain a 3D pointcloud.
Since the pixels of the 2D selection are unnecessarily dense, we uniformly sample 70\% of them; after unprojection, we further compact the pointcloud by removing statistical outliers. In \cref{fig:pc_filtering} we qualitatively show the benefit of filtering the pointcloud.

The pointcloud is rendered using screen-space quads of the same size, and it is blended with the scene using depth testing. The depth testing is performed with the rough depth map obtained from gaussians to guarantee real-time rendering of the two.

The unprojected pointcloud can be refined in multiple rounds by the user. Whenever the user re-enters the selection mode the 3D pointcloud is projected onto the view being edited and the user can possibly clear points as well as add new ones, as it is described in \ref{2d_selection}.

\subsection{Dataset Update and Background Optimization}

While the user has the editor open,
an optimization process runs in the background.
A training camera from the scene dataset is randomly sampled,
the 3D selection -- if any -- is projected onto the camera's image plane and the edit is applied on the projected pixels.

In our work, we restrict possible edit operations to tinting, also known as color filtering or \quotes{recoloring}.
The recoloring operation consists in updating the color of each pixel $c$, marked by the projected mask, to the product of itself and an RGB tuple $t$, as:
\begin{equation}
c^*  = t \cdot c
\end{equation}

Since recoloring does not involve any geometry change, refitting only occurs for the spherical harmonic coefficients, hence no further involvement for gaussians' densification or pruning processes.

As the 3D selection is rendered using screen-space quads, it is not view-consistent across training views, and the adherence of the mask to scene surfaces is left to the user.
The background optimizer is tasked to find the best fit of the colors towards non consistent views.
Since the gaussians are 3D objects, the result will inherently be view-consistent and as we are freezing their geometry, the scene geometry is always preserved.

\section{Implementation Details}

In this section we discuss some implementation details of our real-time editor \url{https://github.com/loryruta/recogs}.
The editor has been implemented in C++ and CUDA using the original 
\href{https://github.com/graphdeco-inria/diff-gaussian-rasterization}{Gaussian Splatting rasterizer}, and its real-time capabilities have been assessed on a NVIDIA RTX 3090.

The editor is split into two logical parts: the visualization and the background optimizer, both running on their dedicated threads. The CUDA kernels dispatched by each run on dedicated CUDA streams, so to maximize GPU throughput and avoid tail effects.

\subsection{Hard-coded Gradient Computation}

We chose not to rely on libtorch in order to eliminate the overhead introduced by the autograd mechanism and to retain greater flexibility for potential optimizations (e.g. kernel fusion).
In Gaussian Splatting autograd is only used for the computation of the final loss gradients, and gradients for the rest of the pipeline are computed by the \href{https://github.com/graphdeco-inria/diff-gaussian-rasterization}{Gaussian Splatting rasterizer} and \href{https://github.com/rahul-goel/fused-ssim}{SSIM} modules. More in detail, we hard-coded the computation of the gradient of the loss w.r.t. the predicted image $\mathbf{y}$ (derivative of \cref{eq:gs_loss}):
\begin{equation}
\begin{gathered}
\dfrac{\partial \mathcal{L}}{\partial \mathbf{y}} =
(1 - \lambda) \dfrac{\partial \mathcal{L}_1}{\partial \mathbf{y}} -
\lambda \dfrac{\partial \mathcal{L}_{SSIM}}{\partial \mathbf{y}} 
\\
\dfrac{\partial \mathcal{L}_1}{\partial y_i}( \cdot ) =
\dfrac{1}{N}
\dfrac{\partial}{\partial y_i}
\left| x \right| =
\dfrac{1}{N}
\sign ( \cdot )
\end{gathered}
\end{equation}
where $y_i$ is each pixel of the predicted image $\mathbf{y} \in \mathbb{R}^N$, and $\partial \mathcal{L}_{SSIM}/{\partial \mathbf{y}}$ is computed by the SSIM module backward.
$\partial \mathcal{L}/{\partial \mathbf{y}}$ is then passed to the rasterizer backward function (\cref{eq:gs_backward}) to provide the gradients of the loss with respect to all scene parameters:
\begin{equation} \label{eq:gs_backward}
\begin{gathered}
\operatorname{Backward}:
\dfrac{\partial \mathcal{L}}{\partial \mathbf{y}}
\to
\left\{
\dfrac{\partial \mathcal{L}}{\partial \mathbf{\mu}},
\dfrac{\partial \mathcal{L}}{\partial \mathbf{R}},
\dfrac{\partial \mathcal{L}}{\partial \mathbf{S}},
\dfrac{\partial \mathcal{L}}{\partial \sigma},
\dfrac{\partial \mathcal{L}}{\partial \mathbf{C}_l^m},
\dfrac{\partial \mathcal{L}}{\partial \mathbf{\mu}''}
\right\}
\end{gathered}
\end{equation}
of which we only consider the gradients of the loss w.r.t. spherical harmonics coefficients $\partial \mathcal{L}/\partial \mathbf{C}_l^m$, that are used by the optimizer to step their value towards the edited dataset.

\subsection{Use HWC for Rasterization Output}

For better GPU memory efficiency, we found advantageous to use the HWC format for the rasterizer output and subsequently transition it to CHW for computing the SSIM loss, that is convolution based. Profiling revealed that the \texttt{renderCUDA} kernel’s L1 cache hit rate increased from 4.12\% to 55.70\%, thereby reducing its execution time from 6.61 ms to 1.52 ms.

\subsection{Optimizing PCVNet Inference}

\begin{table}
\centering
\begin{tabular}{lrr} 
 \toprule
 Method & Inference & Model \\
  & speed (s) & size (Mb) \\
 \midrule
 PCVNet (PyTorch) & 0.71 & 50 \\
 PCVNet (CUDA) & 1.87 & 50 \\
 PCVNet (TensorRT) & 0.49 & 102 \\
 PCVNet (TensorRT+FP16) & 0.20 & 22 \\
 \bottomrule
\end{tabular}
\caption{Inference speed and model size comparison. 
The model size is obtained from the checkpoint file size, the .onnx file size and the engine for PyTorch, ONNX and TensorRT respectively.
}
\label{table:pcvnet_runtime}
\end{table}

We focused on reducing the depth estimation inference time, as well as the model size, while retaining acceptable depth maps.
Lacking of ground truth depth maps on the scenes we tested, we evaluated the accuracy of the depth maps qualitatively by visualizing the pointcloud obtained after unprojection.
Our ideal objective was to achieve real-time performance (i.e., under 33 ms) as to use the depth estimation model consistently in our interactive editor, thereby avoiding reliance on the approximate depth derived from gaussians.

\begin{figure}[h!]
  \centering
  \includegraphics[width=0.9\linewidth]{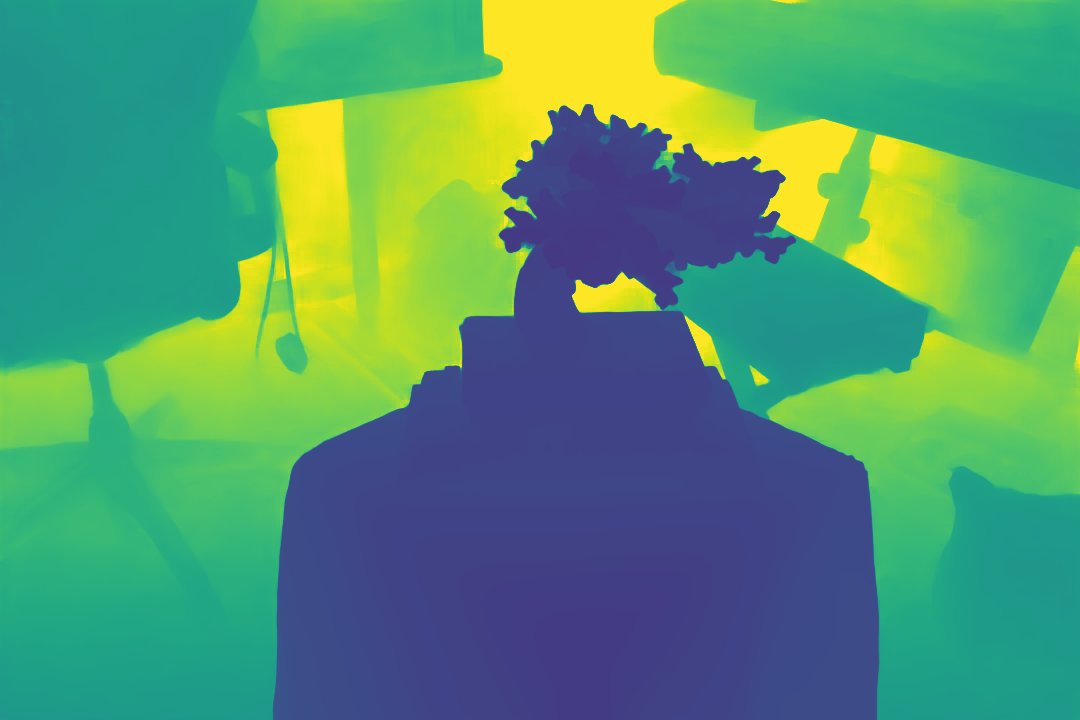}
  \includegraphics[width=0.9\linewidth]{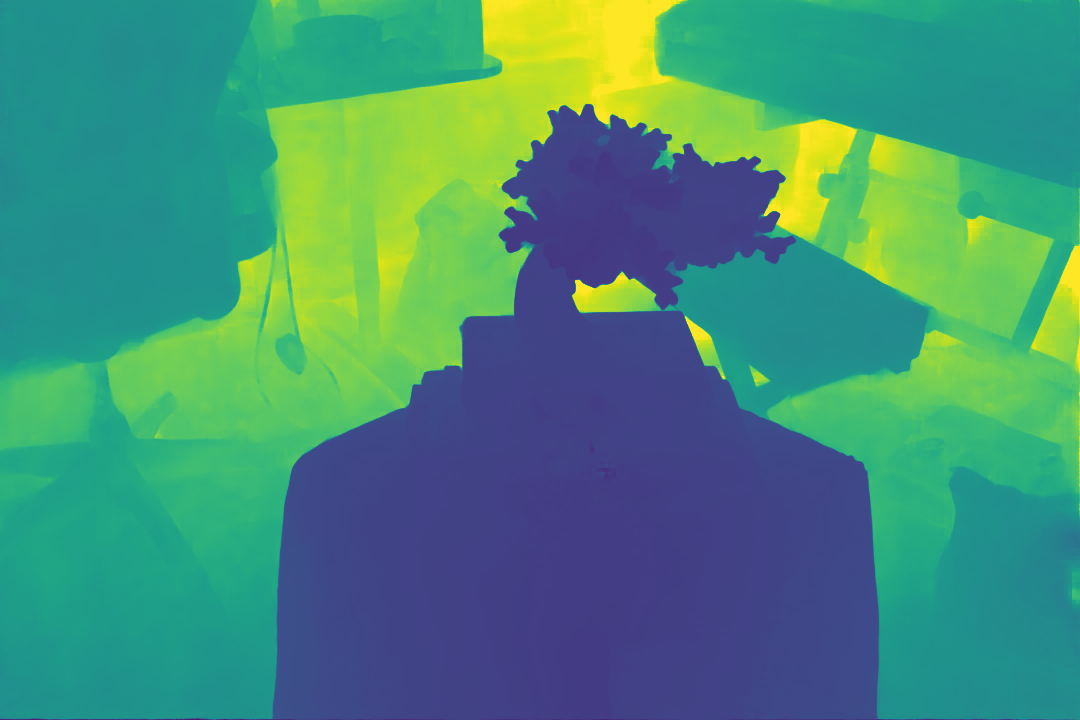}
  \caption{
  Top image is PCVNet prediction from libtorch.
  Bottom image is the prediction from the TensorRT engine, with FP16 enabled. Although the latter is more noisy, the unprojected pointcloud after outlier filtering is still acceptable.
  }
  \label{fig:pcvnet_quant_depthmap_comparison}
\end{figure}

The combination of PCVNet using TensorRT and FP16 precision achieved the best runtime and model size (see \cref{table:pcvnet_runtime} for more details), yet not enough for per-frame inference. The degradation brought by the optimized model can be evinced by  \cref{fig:pcvnet_quant_depthmap_comparison}.  

\section{Evaluation}

To the best of our knowledge, none of the previous works propose a selection nor editing approach like ours. 

In \cref{table:feature_comparison}, we provide a brief taxonomy of the different methods we have investigated. Those that are diffusion -based \cite{tip_editor}, \cite{vcedit}, \cite{gaussctrl}, \cite{dge}, \cite{editsplat} are also text-prompted, a type of input which prohibits pixel-level granularity.
As an advantage, as these approaches rely on the rich priors of pre-trained diffusion model, they can enable complex edits, possibly involving geometry modifications; however this has the cost of hindering view-consistency.

Most notably, contemporary diffusion models are resource intensive and slow at inference. To support our claims, we have selected and benchmarked DGE \cite{dge} on our hardware \footnote{A RTX NVIDIA 3060, supported by 64 GB RAM and a i7-11700KF.} to edit diverse scenes. We have chosen DGE beacause it was compared with several editing approaches, is designed to solve view-consistent editing, and code is available.
Results show $>20$ Gb peak memory and 4 minutes on average to execute the pipeline.

\begin{figure}[h!]
  \centering
  \includegraphics[width=\linewidth]{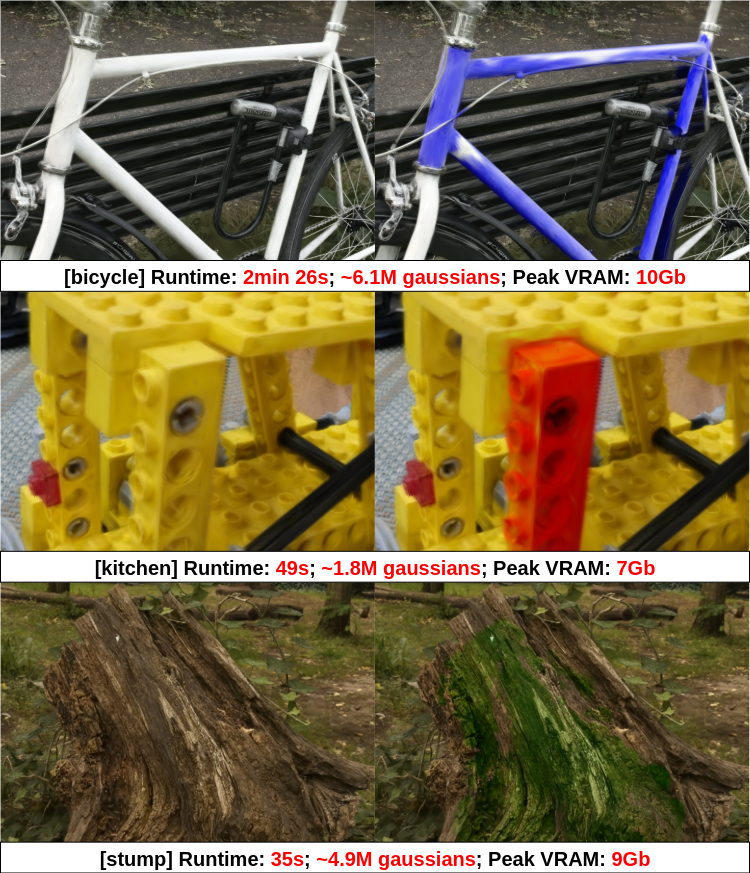}
  \caption{Elapsed time and GPU memory consumption to perform edits on 3 MipNeRF360 scenes: bicycle, kitchen, and stump.}
  \label{fig:recogs_runtimes}
\end{figure}

In \cref{fig:recogs_runtimes} we showcase the runtime for applying recoloring on 3 diverse scenes with the editor integrating our pipeline.
The time reported is the elapsed time from when the user finishes the selection, to the point the right screenshot was taken (the optimizer could continue running). The runtime is proportional to the scene size, and is higher if the edited region is infrequent in the training set.
The bicycle scene has the highest runtime both because it is the heaviest, and because the edit was applied on an infrequent area (many training views look at the front of the bench).
On the other hand, the stump scene is heavy as well but the edit was applied on a frequent region.
Regarding memory, GPU memory usage peaked 10 GB for the bicycle scene.

Ultimately, PaintSplat \cite{paintsplat} proposes to address our same goal of real-time interactive editing. Their editing approach is to paint 3D stamps onto a splat. This could possibly address re-coloring; however, code is not yet released, so we could not properly compare it. Regardless, we still find our proposed approach as an alternative interactive editing method.


\begin{table*}[tbh!]
\centering
\begin{tabular}{lccccc}
\toprule
\makecell{Method} &
\makecell{Family} &
\makecell{View Cons.} &
\makecell{Selection type} &
\makecell{Edit type} &
\makecell{Code} \\
\midrule
TIP-Editor & 2D Diff     & N & Text, Image, AABB & Any & Y \\
VcEdit     & 2D Diff     & Y & Text              & Any & Y \\
GaussCtrl  & 2D Diff     & Y & Text              & Any & Y \\
DGE        & 2D Diff     & Y & Text              & Any & Y \\
EditSplat  & 2D Diff     & Y & Text              & Any & Y \\
ICE-G      & SAM         & Y & Text              & Texture, Color & N \\
PaintSplat & Splat Brush & Y & Manual            & Texture, Color, Place Objects & N \\
ReCoGS (ours) & Pixel Brush & Y & Manual & Color & Y \\
\bottomrule
\end{tabular}
\caption{
Comparison of the type and supported features of the existing methods for Gaussian Splatting editing. 
}
\label{table:feature_comparison}
\end{table*}



\section{Limitations and Future Work}

Our method presents imperfections coming from errors of the depth estimation model, the pointcloud not being view-consistent (i.e., not in world-space), and simplifications done to suit the pipeline to interactive execution. Given the great participation of the user during the editing process, we believe that many of them could be mitigated through minimal additional user interaction.

\begin{figure}[!tb]
  \centering
  \includegraphics[width=\linewidth]{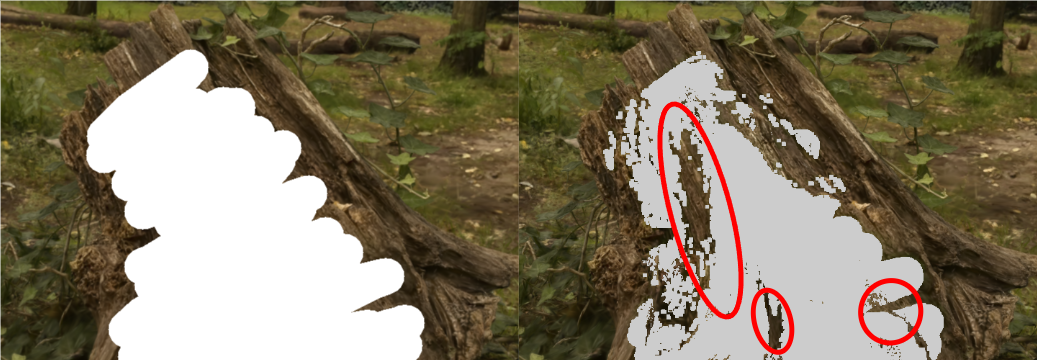}
  \caption{The unprojected 3D selection might not render as a continuous surface. The \quotes{holes} shown in this figure are either caused by errors in depth prediction or by the fact that - to perform depth testing between the scene and the 3D selection - we use the imprecise depth map obtained from gaussians to ensure real-time performance.}
  \label{fig:holes_in_selection}
\end{figure}

As shown in \cref{fig:holes_in_selection}, after unprojection the 3D selection could present holes. Those are caused both by errors in depth prediction and by the fact that we adopt the imprecise \quotes{depth from gaussians} to perform depth testing in real-time when rendering the 3D selection and the scene together.

Furthermore, as we only optimize spherical harmonics, the edit performed by the user might leak whether the gaussian subject to the edit is bigger than the selection. Segmentation and re-enabling densification on the selected areas can be explored to address the issue. Part of future work could also target editing speed: each training camera could be assigned with an importance proportional to how much influence it has on the edit.

Finally, while PCVNet \cite{pcvnet} offered a good accuracy/efficiency trade-off, we believe that tailoring a stereo matching model to gaussian splatting, for example by seeding it with gaussians' depth, may yield greater improvements.

\section{Conclusion}

In this article, we have presented a novel approach for selecting and editing a 3D reconstruction capable to run on a customer -level hardware, preserving real-time interactivity. We have put effort in realizing our pipeline in a software featuring several optimizations to guarantee the best performance.
As a simple edit we focused on recoloring, but we believe our \quotes{2D to 3D} selection method to possibly be a driving concept or inspiration for more complex selections and edits.

\clearpage
\newpage

\bibliographystyle{eg-alpha-doi} 
\bibliography{main}

\end{document}